\documentclass[conference]{IEEEtran}
\IEEEoverridecommandlockouts
\usepackage{cite}
\usepackage{amsmath,amssymb,amsfonts}
\usepackage{algorithmic}
\usepackage{graphicx}
\usepackage{textcomp}
\usepackage{xcolor}
\usepackage{booktabs}
\usepackage{tabularx}
\usepackage{geometry}
\usepackage{adjustbox}
\usepackage{float}
\usepackage{multirow}
\usepackage{hyperref}
\def\BibTeX{{\rm B\kern-.05em{\sc i\kern-.025em b}\kern-.08em
    T\kern-.1667em\lower.7ex\hbox{E}\kern-.125emX}}
\begin{document}

\title{Liaohe-CobotMagic-PnP: an Imitation Learning Dataset of Intelligent Robot for Industrial Applications \\
\thanks{This work has been submitted to the IEEE for possible publication. Copyright may be transferred without notice, after which this version may no longer be accessible.This work is supported by Liaoning Liaohe Lab with project number LLL24ZZ-02-01 and LLL24ZZ-02-02.}
}
\author{
    \IEEEauthorblockN{
        Yizhe Chen\IEEEauthorrefmark{2},
        Qi Wang\IEEEauthorrefmark{2},
        Dongxiao Hu\IEEEauthorrefmark{3},
        Jingzhe Fang\IEEEauthorrefmark{4},\\
        Sichao Liu\IEEEauthorrefmark{6},
        Zixin An\IEEEauthorrefmark{5},
        Haoran Liu\IEEEauthorrefmark{4},
        Hongliang Niu\IEEEauthorrefmark{4},\\
        Dong Li\IEEEauthorrefmark{8},
        Chuanfeng Feng\IEEEauthorrefmark{2}\textsuperscript{*},
        Dapeng Lan\IEEEauthorrefmark{9},\textsuperscript{*}
        Yu Liu\IEEEauthorrefmark{9},
        Zhibo Pang\IEEEauthorrefmark{6}
    }\\

    \IEEEauthorblockA{\IEEEauthorrefmark{2}School of Information Science and Engineering, Shandong Normal University, Shandong, China\\
    2023317064@stu.sdnu.edu.cn, fengcf@sdnu.edu.cn, wangqi@sdnu.edu.cn}

    \IEEEauthorblockA{\IEEEauthorrefmark{3}School of Advanced Technology, Xi’an Jiaotong-Liverpool University, Jiangsu, China\\
    dongxiao.hu22@student.xjtlu.edu.cn}

    \IEEEauthorblockA{\IEEEauthorrefmark{4}Liaohe Laboratory, Innovation Institute of Intelligent Robotics Shenyang Co., Ltd., Shenyang, China\\
    fangjingzhe@niir.com.cn, liuhaoran@niir.com.cn, niuhongliang@niir.com.cn}

    \IEEEauthorblockA{\IEEEauthorrefmark{5}School of Information Engineering, Shenyang University of Chemical Technology, Shenyang, China\\
    771112020@qq.com}

    \IEEEauthorblockA{\IEEEauthorrefmark{6}
    Department of Production Engineering \textit{and} Department of Intelligent Systems,\\
    KTH Royal Institute of Technology, Stockholm, Sweden\\
    sicliu@kth.se, zhibo@kth.se}

    \IEEEauthorblockA{\IEEEauthorrefmark{8}Shenyang Institute of Automation, Chinese Academy of Sciences, Shenyang, China\\
    lidong@sia.cn}

    \IEEEauthorblockA{\IEEEauthorrefmark{9}
    University of Chinese Academy of Sciences, Beijing, China\\
    dapengl@ieee.org, liu.yu@ieee.org}

    \thanks{\textsuperscript{*}Corresponding authors: Dapeng Lan (dapengl@ieee.org), Chuanfeng Feng (fengcf@sdnu.edu.cn).}
}

\maketitle

\begin{abstract}
In Industry 4.0 applications, dynamic environmental interference induces highly nonlinear and strongly coupled interactions between the environmental state and robotic behavior. Effectively representing dynamic environmental states through multimodal sensor data fusion remains a critical challenge in current robotic datasets. To address this, an industrial-grade multimodal interference dataset is presented, designed for robotic perception and control under complex conditions. The dataset integrates multi-dimensional interference features including size, color, and lighting variations, and employs high-precision sensors to synchronously collect visual, torque, and joint-state measurements. Scenarios with geometric similarity exceeding 85\% and standardized lighting gradients are included to ensure real-world representativeness. Microsecond-level time-synchronization and vibration-resistant data acquisition protocols, implemented via the Robot Operating System (ROS), guarantee temporal and operational fidelity. Experimental results demonstrate that the dataset enhances model validation robustness and improves robotic operational stability in dynamic, interference-rich environments. The dataset is publicly available at: \href{https://modelscope.cn/datasets/Liaohe_LAB/Liaohe-CobotMagic-PnP/files}{https://modelscope.cn/datasets/Liaoh\_LAB/Liaohe-CobotMagic-PnP}.
\end{abstract}


\section{Introduction}

Existing robot dataset construction has emphasized scene diversity and task coverage by integrating multimodal environmental data across domains like domestic service, industrial manufacturing, and office settings. These datasets incorporate multi-stage task workflows, multimodal perception integration, and dynamic interaction modeling to support cross-scenario generalization. However, current frameworks show limitations in an aspect: they inadequately model complex intra-scene dynamic characteristics and do not fully leverage the deep correlations between scene elements~\cite{andry2001learning}, thus underutilizing the synergistic potential of multimodal spatiotemporal information.

Current datasets face limitations in modeling industrial multimodal disturbances, particularly in lighting conditions, interference object shapes, and color representation. For lighting, existing datasets lack comprehensive coverage of complex variations such as high dynamic range illumination. Industrial environments typically require illuminance levels exceeding 300 lx, with precision tasks demanding 500–1000 lx, while residential settings show lower requirements—100 lx for living rooms versus 150 lx for kitchen workspaces. This disparity reduces robotic adaptability to dynamic industrial lighting scenarios. In terms of object shapes, datasets primarily feature simplistic geometries, inadequately representing the irregular forms and multi-scale sizes common in industrial contexts, which compromises navigation accuracy. Regarding colors, insufficient diversity in color modeling persists, especially in scenarios where interference objects share chromatic similarity with targets or exhibit non-standard spectral characteristics. These gaps may lead to reduced discrimination accuracy in real-world deployments.

The Liaohe-CobotMagic-PnP dataset addresses the critical challenges of perception and control in dual-arm cooperative robotic systems operating within dynamic Industry 4.0 environments. Designed to mitigate the limitations of existing datasets in modeling industrial-grade multimodal interferences, this dataset integrates gradient variations in illumination conditions, positional displacement interference, geometrically similar interfering objects, and color-based ambiguities between target and distractor objects. This paper demonstrates the reliability and effectiveness of the proposed industrial multi-modal interference dataset in complex scenarios through both hardware testing and model verification. At the hardware level, by leveraging the ROS system, it achieves microsecond-level timestamp synchronization and dynamic anti-vibration interference capability. The release frequency of joint status topic messages remains stable, satisfying the timeliness demands of industrial control. In model verification, the Action Chunking with Transformers (ACT) model attains a success rate of more than 60\% in dynamic jamming tasks, highlighting the agility of lightweight models. Meanwhile, through multimodal pretraining and hierarchical task decomposition, the large Vision-Language-Action Flow Model for General Robot Control ($\pi0$) reaches a task success rate greater than 40\%, which is markedly superior to conventional methods. Furthermore, the dataset enables multitask hybrid training and cross-scale model optimization. It offers a unified benchmark for enhancing the high-precision and robust control of industrial robots and drives the development of industrial automation toward an integration of perception and decision-making based on virtual reality fusion.

\section{Related Works}

\begin{table*}[!htb]
\centering
\Large
\renewcommand{\arraystretch}{1.5} 
\setlength{\extrarowheight}{2pt}   

\caption{Comparison of Robot Dataset: Multimodal Characteristics, Scene Limitations, and Industrial Adaptability Analysis}
\label{tab:framework_comparison}
\begin{adjustbox}{max width=\textwidth} 
\begin{tabular}{@{}lllll@{}} 
\toprule
Publisher/Organization & Scale \& Features & Application Scenarios & Core Data Characteristics & Industrial Scene Limitations \\
\midrule
Google DeepMind \& 21 orgs & 60 datasets (22 morphologies, 1M trajectories) & Home/industrial ops & Multimodal (RGB/depth/action) + RLDS & Limited to controlled lab environments, lacking industrial noise and dynamic disturbances \\
AgiBot & 1M+ real-world data (3K objects) & Complex industrial tasks & Long-horizon planning & Requires extensive fine-tuning for real-time feedback \\
Tsinghua TSAIL & 1.2B param model (300 tasks) & Dual-arm manipulation & Multimodal RL & Absence of torque parameters and joint velocity data \\
UC Berkeley BAIR & 15M video frames (7 platforms) & Visual model & Multi-view RGB & No control commands or industrial context \\
Offline RL Benchmark & 20+ tasks subsets & Policy optimization & State-action pairs & Lacks high-frequency sensor data and industrial task domains \\
Stanford & 8k expert demos (cleaning/cooking) & Task planning & Language+vision fusion & No industrial skill requirements or disturbance modeling \\
CMU Robotics & 6D pose estimation (21 objects) & Object manipulation & High-precision annotations & Static lab environments without dynamic interference \\
\bottomrule
\end{tabular}
\end{adjustbox}
\end{table*}


In recent years, the field of robotics learning has significantly promoted technological breakthroughs such as cross-platform adaptation, complex decision-making, and fine control by building diverse large-scale data sets, laying the foundation for general robotics capabilities in real scenarios. In particular, Open X-Embodiment integrates datasets from 21 institutions worldwide~\cite{vuong2023open}, encompassing 22 different robotic platforms and over 160,000 tasks. This large-scale effort supports cross-robot generalization and provides a foundation for multiscenario transfer learning. AgiBot World~\cite{bu2025agibot} offers millions of real-world robot interaction samples across more than 100 scenarios - including domestic, industrial, and service domains - demonstrating robust performance in long-term dynamic task planning. RDT showcases dual-arm manipulation and generalization to previously unseen objects through reinforcement learning over 300+ task demonstrations. D4RL~\cite{fu2020d4rl} contributes a comprehensive benchmark for offline policy optimization, facilitating research on complex state-action space exploration. The ALFRED dataset~\cite{yao2023react} integrates natural language instructions with visual observations to enable long-horizon planning for household tasks. Meanwhile, the YCB Video dataset~\cite{grenzdorffer2020ycb} provides high-precision RGB-D annotations, establishing a widely adopted benchmark for grasping objects and estimating 6D poses. Collectively, these datasets have accelerated the development of robotic perception, decision making, and control systems, as summarized in Table~\ref{tab:framework_comparison}.

Although existing datasets have shown excellent performance in general robotics applications, they are still insufficient in modeling industrial-level multimodal interference and cannot capture the differences between industrial scenarios and other scenarios. This study centers on the creation of industrial-grade multimodal interference modeling, introducing a new dataset that integrates various industrial-grade interference elements such as size, color, and lighting. Crafted to address the challenges of complexity and dynamic interference in industrial environments, this dataset provides a more comprehensive and realistic representation of robot-related industrial scenarios. The research ultimately aims to establish a robust foundation for enhancing industrial robots' adaptability and resilience in demanding settings and to foster the integration of perception and decision-making in industrial automation.

\section{Experimental Environment}\label{AA}
\subsection{Scene Description}
A customized industrial simulation platform was deployed within the operational area of the Cobot Magic dual arm at Agilex Robotics~\cite{SonglingRobotics2022}. The workspace configuration comprises an antistatic white workbench sized 1800 × 600 $\mathrm{mm}^2$, positioned at a height of 600 mm to align with the robot base, thus guaranteeing consistent operations. The objects designed for manipulation are metallic cubes in red, blue, and green, each featuring a matte surface finish. This surface treatment is intended to minimize specular reflections that could interfere with visual sensing mechanisms. Furthermore, two varieties of objects necessitating helium gas detection are arbitrarily placed on the workbench. This setup is designed to replicate the variable task conditions that might be encountered in a real industrial setting.

\begin{figure}[htbp]  
  \centering          
  \includegraphics[width=0.5\textwidth]{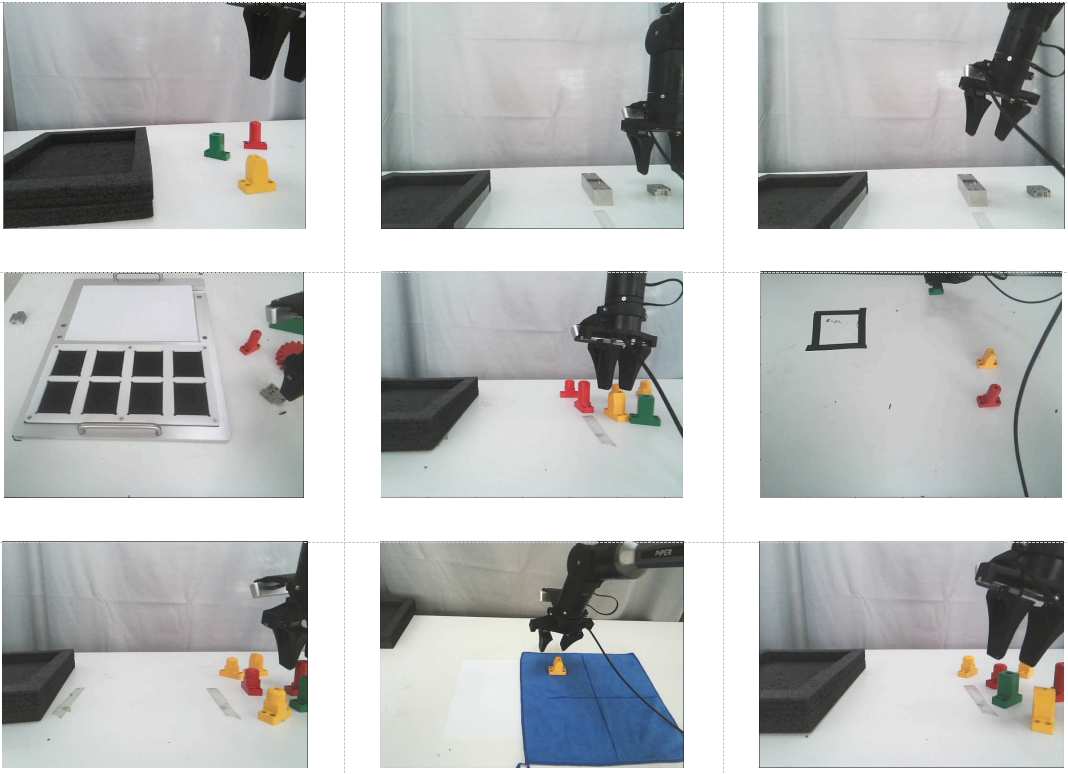}  
  \caption{Liaohe-CobotMagic-PnP dataset: Example of pick and place tasks in multiple interference scenarios}  
  \label{fig:changjing} 
\end{figure}

To mimic realistic industrial environments, non-essential obstacles have been systematically removed from the workspace. In addition, the lighting conditions within the platform are alternated between natural and factory lighting regimes. This approach aims to evaluate the performance of the models under various visual inputs. For factory lighting conditions, an illumination intensity ranging from 750 to 300 Lux is adopted. This specific range of illumination is selected to reflect the common lighting conditions found in typical industrial factory settings, providing a practical and relevant test scenario to assess the robustness and adaptability of robotic systems in such environments. The scenario in the dataset is shown in Fig.~\ref{fig:changjing}.

During the data collection process, a six-axis motion platform introduces random displacements with ±0.5 mm amplitude to simulate industrial vibrations. Interference materials with geometric similarity exceeding 85\% to the target object, including variations in the frosted or polished surface, are injected simultaneously. A constant current-driven LED array ensures lighting stability, maintaining color temperature fluctuation within ±2\%. These operations implicitly embed real industrial scene variables into the dataset.


  

\subsection{Hardware configuration}

\begin{figure}[htbp]  
  \centering          
  \includegraphics[width=0.5\textwidth]{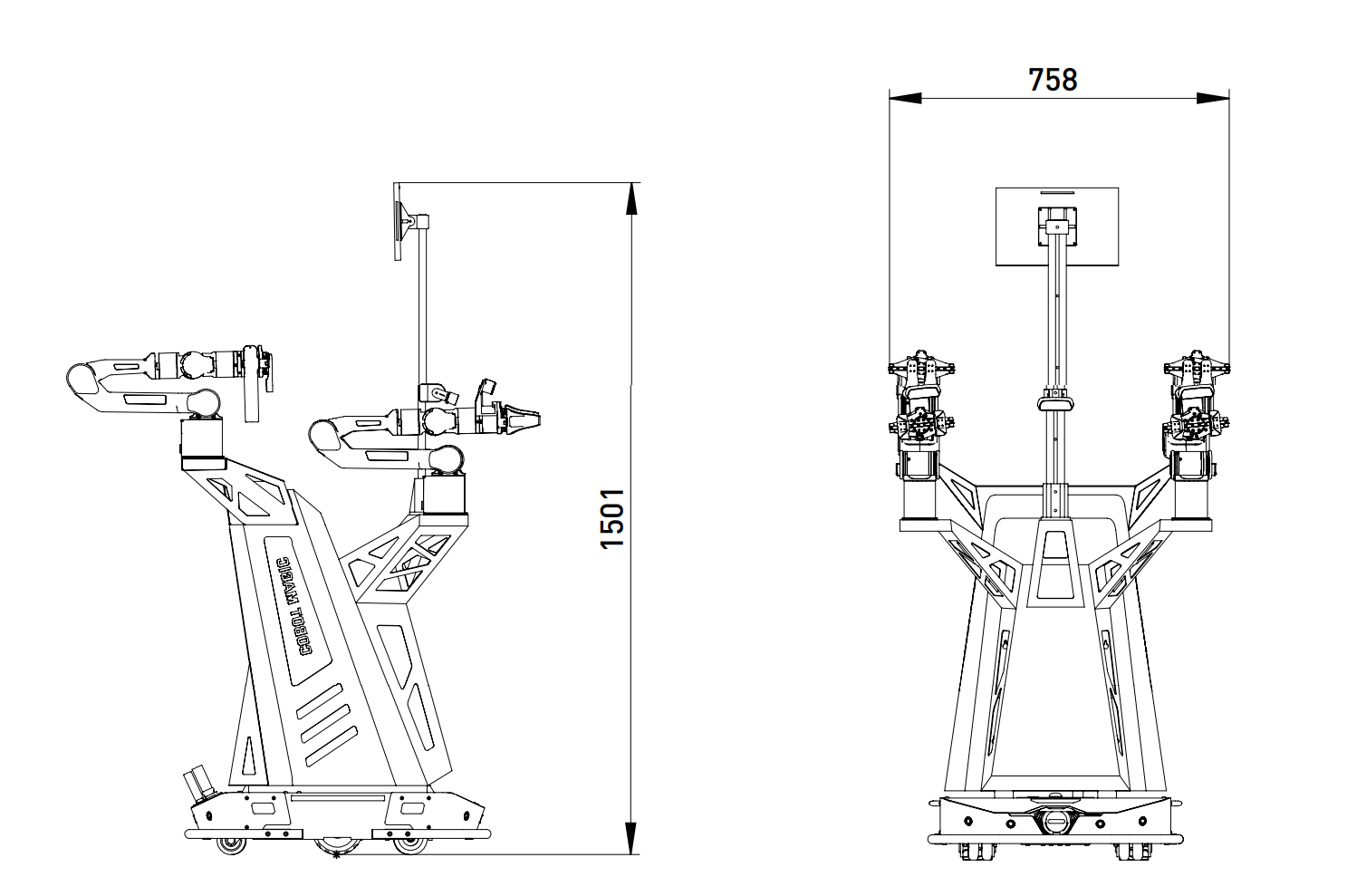}  
  \caption{3D image of robot controlled by Agilex's Cobot Magic system\cite{cobotmagic_2025}}  
  \label{fig:3d} 
\end{figure}

The Cobot Magic platform developed by Agilex Robotics has been extensively adopted in research due to its high reliability and operational versatility~\cite{mu2024robotwin}. As depicted in Fig.~\ref{fig:3d}, the robot played a pivotal role in the ManiBox project, a collaborative effort between Tsinghua University and the National University of Singapore~\cite{liu2024rdt}, where it successfully executed complex grasping and manipulation tasks. The stable dual-arm configuration of the system demonstrated strong adaptability and precision in both simulated and real-world environments. Furthermore, in the JD ManiData dataset, Cobot Magic was employed to perform a broad range of manipulation tasks, exhibiting high task success rates and consistent motion execution. These empirical results collectively affirm the suitability of the platform for advanced research in dual-arm robotic systems and underscore its potential for further development in industrial and academic applications.

\section{Dataset Structure}

\begin{table*}[htbp]
\centering
\caption{Task types of Liaohe-CobotMagic-PnP dataset: Construction and Scale Analysis of Multi-Interference Scenarios for Capturing and Placement Tasks}
\setlength{\extrarowheight}{1pt}
\label{tab:dataset_details}
\begin{tabularx}{\textwidth}{@{}c@{\ } >{\raggedright\arraybackslash}p{4cm}@{\ }c@{\ }c@{\ }c@{}}  
\toprule
\textbf{Task Category}         & \textbf{Task Description}                                                                 & \textbf{Disturbance Items}                & \textbf{Dataset Count} & \textbf{Light Condition} \\  
\midrule
\multirow{4}{*}{Blue Tablecloth Area Tasks} 
                            & Grasp yellow workpiece from Area 11                                                   & None                            & 24 (Success) + 16 (Fail) & Natural \\  
                            & Grasp yellow workpiece from Area 12                                                   & None                            & 24 (Success) + 16 (Fail) & Natural \\
                            & Grasp yellow workpiece from Area 21                                                   & None                            & 24 (Success) + 16 (Fail) & Natural \\
                            & Grasp yellow workpiece from Area 22                                                   & None                            & 24 (Success) + 16 (Fail) & Natural \\
\midrule
Workpiece Placement Task 1  & Place green workpiece into black frame area                                             & Red/Yellow workpieces             & 180                & Natural \\
Workpiece Placement Task 2  & Insert large iron block into black box                                                  & Small iron block                  & 30                 & Natural \\
Workpiece Placement Task 3  & Insert small iron block into black box                                                  & Large iron block                  & 30                 & Natural \\
Workpiece Placement Task 4  & Insert large iron block into black box                                                  & None                            & 958                & Factory Lighting \\  
Workpiece Placement Task 5  & Place green workpiece into black box                                                    & Multiple color workpieces         & 40                 & Natural \\  
Workpiece Placement Task 6  & Place yellow workpiece into black box                                                   & Multiple color workpieces         & 40                 & Natural \\  
Workpiece Placement Task 7  & Place red workpiece into black box                                                      & Multiple color workpieces         & 40                 & Natural \\
\midrule
\multirow{4}{*}{Special Placement Tasks}
                            & Place yellow workpiece into red area (Retry on failure)                               & 2 yellow workpieces               & 30                 & Natural \\
                            & Place yellow workpiece into red area (Standard placement)                             & 2 yellow workpieces               & 30                 & Natural \\
                            & Place yellow workpiece into red area (Single interference)                            & 1 yellow workpiece                & 30                 & Natural \\
                            & Place yellow workpiece into red area (No interference)                                & None                            & 30                 & Natural \\
\midrule
Special Placement Task 5    & Place raw material into vise clamp                                                    & None                            & 180                & Factory Lighting \\  
Special Placement Task 6    & Insert small iron block into black testing area (Helium detection)                      & None                            & 392                & Factory Lighting \\  
\bottomrule
\end{tabularx}
\end{table*}

\begin{figure}[htbp]  
  \centering          
  \includegraphics[width=0.5\textwidth]{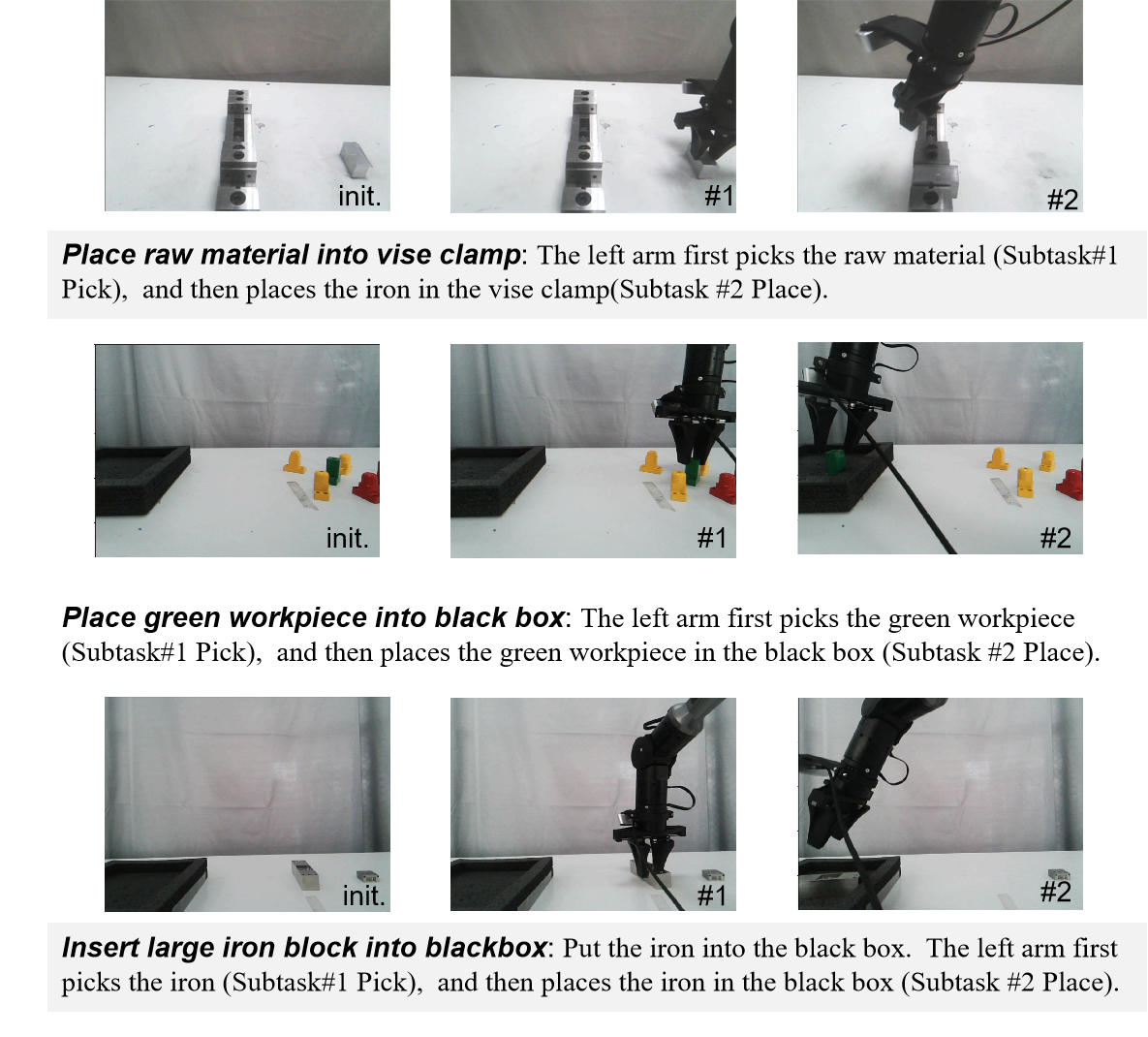}  
  \caption{Dataset task example: Give track information of some tasks.}  
  \label{fig:guijishili} 
\end{figure}
\begin{figure*}

  \centering          
  \includegraphics[width=6.5in]{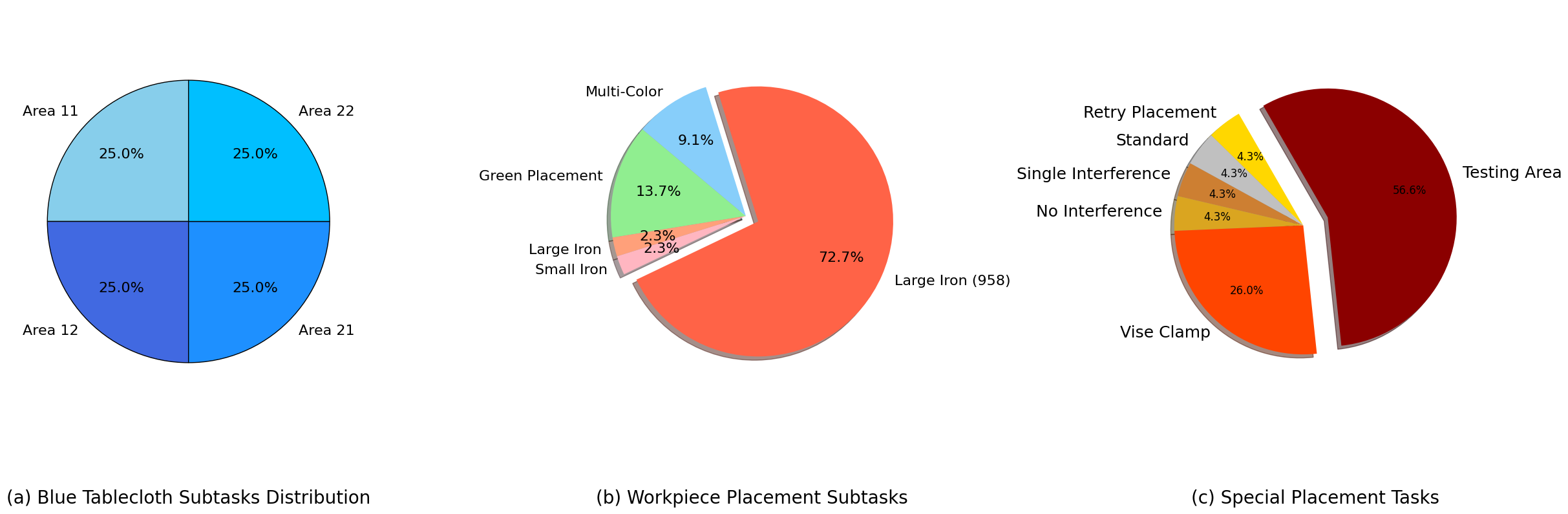}  
  \caption{Proportion of Liaohe-CobotMagic-PnP datasets of various types: (a) The blue tablecloth operation tasks are evenly distributed in four areas; (b) The work piece placement task is mainly focused on the processing of large iron pieces (958), accounting for ($72.7\%$; (c) In special placement tasks, the test area occupies an absolute dominant position (($96\%$), while others such as fixture fixation, interference free and single interference tasks occupy a relatively small proportion.}  
  \label{fig:1} 
  
\end{figure*}

The proposed dataset is designed to support imitation learning for industrial robotic applications, encompassing three core task categories: object capture, placement, and detection. It comprises a total of 2,320 multimodal samples, of which approximately 18.5\% represent failure cases, thus enriching the utility of the dataset for robust model training. As illustrated in Fig.~\ref{fig:1}, the dataset simulates the complexity of the production line through a dual encoding scheme based on color regions and a gradient design of dynamic disturbances. The dataset includes a wide range of task complexities, spanning basic operations such as directional placement of green and yellow workpieces, advanced interactions such as area number recognition and failure recovery, and sensor-fusion challenges such as helium gas detection and visual-tactile alignment. To support spatial awareness, the workspace is divided into four distinct zones using a blue tablecloth, enhancing the model’s ability to learn area boundary discrimination. Furthermore, the dataset includes 392 helium detection instances, which promote cross-modal learning by linking physical properties with semantic labels. A detailed summary of the components of the dataset is provided in Table~\ref{tab:dataset_details}.

To meet the robustness requirements of industrial applications, the proposed dataset introduces a novel third-order interference gradient mechanism, which incrementally increases task complexity across three levels: no interference, monochromatic interference, and mixed multi-color interference. A total of 432 failure instances, such as object slippage and misclassification, are annotated with precise timestamps to facilitate the development and training of error correction strategies. At the data acquisition level, high-resolution spatio-temporal annotation is achieved through the integration of an RGB camera with 640 × 480 resolution at 30 fps and six-axis force sensors. Robotic actions are systematically decomposed into a four-stage operational pipeline comprising positioning, planning, control, and verification, with pose errors maintained within a tolerance of ±0.1 mm. In addition, a bidirectional mapping framework is established between the global coordinate system and the local coordinate frame of the piece of work, ensuring compatibility with robot kinematic modeling and improving the applicability of the data set to real-world deployment scenarios. An example of the task trajectory in the data set is shown in Fig.~\ref{fig:guijishili}.

At the application level, the dataset is designed to support two representative industrial scenarios: quality inspection and intelligent warehousing. It enables precise sorting of components, such as helium-based detection of small metallic objects, as well as subpackaging of multicategory mixtures. The inclusion of 392 annotated detection tasks under interference conditions enhances the robustness and classification accuracy of the model in complex operational environments. Furthermore, the data set incorporates a black-box/area task framework that interfaces directly with Warehouse Management System (WMS) instructions. Through the implementation of an area numbering system, the dataset facilitates efficient alignment between semantic directives and physical workspace locations, thereby supporting high-level task abstraction and spatial reasoning in automated warehousing operations.

\section{Technical Verification }
\subsection{Hardware performance }
In ROS communication systems~\cite{koubaa2017robot}, the alignment of timestamps and the stability of the topic message sending frequency are identified as two fundamental mechanisms that play a crucial role in ensuring the reliability of the data.

Regarding timestamp alignment, ROS offers a variety of synchronization strategies via the message\_filters library, such as TimestampSynchronizer and ApproximateTimeSynchronizer. These strategies enable millisecond or subsecond alignment based on the timestamps of messages, effectively addressing timing misalignment issues that arise from transmission delays or differences in sampling rates during the fusion of multisensor data. For example, it is feasible to achieve synchronization of IMU data within a 0.1-second error range by using the approximateTimePolicy, thus maintaining temporal consistency~\cite{breiling2017secure}.

In terms of the frequency stability of the sending of messages, the Quality of Service (\text{QoS}) configurations in ROS are vitally important. By configuring \text{ReliabilityPolicy} (for example, employing the RELIABLE mode to enable message retransmission) and \text{HistoryPolicy} (e.g., utilizing KEEP\_LAST to control the queue depth), in conjunction with precise timer scheduling, the predefined publishing frequency can be effectively maintained.

Furthermore, ROS distributed time management provides support for physical clock synchronization protocols, such as the Precision Time Protocol (PTP), which effectively eliminates clock drift between devices. This offers fundamental timing support for high-frequency data streams. Collectively, these mechanisms enable ROS to achieve high-reliability, low-latency data communication in perception and control applications.

\paragraph {\textbf{ROS topic timestamp alignment}}
\begin{figure*}[htbp]  

  \centering          
  \includegraphics[width=0.8\textwidth]{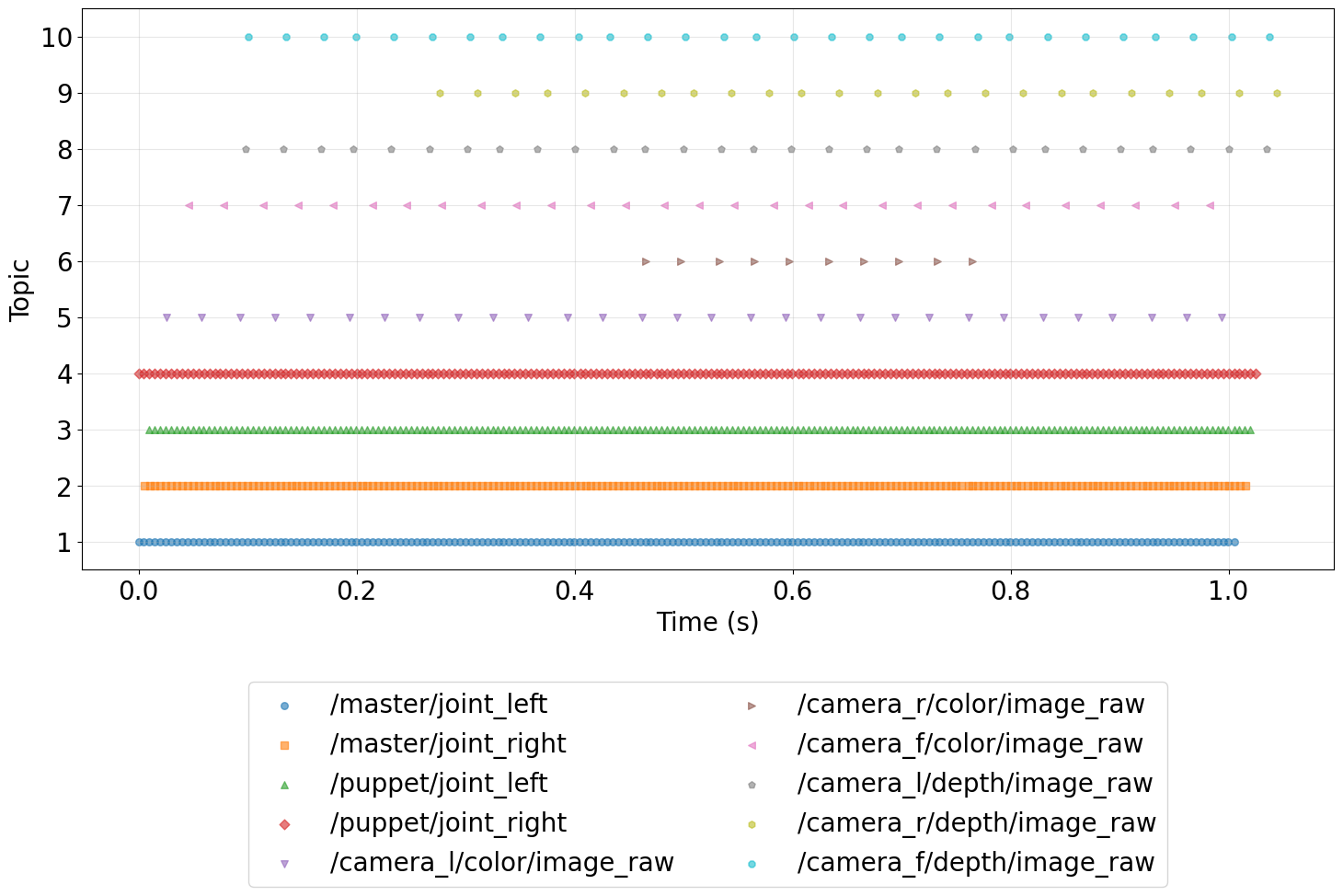}  
  \caption{ROS topic timeline diagram}  
  \label{fig:shijianchuo} 
  
\end{figure*}
\begin{table}
\centering
\footnotesize 
\setlength{\tabcolsep}{1pt} 
\caption{ROS Node Communication Delay Distribution: Mean and Standard Deviation Comparison of Multi-Sensor Themes. }
\label{table:my_table}
\begin{tabularx}{1\columnwidth}{p{0.5\columnwidth} llll}
\toprule
\textbf{Topic Name} & \textbf{Message} & \textbf{Mean(ms)} & \textbf{Std(ms)} \\
\midrule
/master/joint\_left & 196 & 5.00 & 0.01 \\
/master/joint\_right & 202 & 5.00 & 0.02 \\
/puppet/joint\_left & 203 & 5.00 & 0.01 \\
/puppet/joint\_right & 202 & 5.00 & 0.01 \\
\midrule
/camera\_l/color/image\_raw & 25 & 40.50 & 13.43 \\
/camera\_r/color/image\_raw & 23 & 45.63 & 15.53 \\
/camera\_f/color/image\_raw & 25 & 41.83 & 15.19 \\
/camera\_l/depth/image\_raw & 31 & 33.36 & 2.49 \\
/camera\_r/depth/image\_raw & 30 & 33.34 & 2.50 \\
/camera\_f/depth/image\_raw & 30 & 33.32 & 2.53 \\
\bottomrule
\end{tabularx}
\end{table}
As depicted in the time sequence diagram~\cite{li2022worst}, the timestamp sequences of the four topics almost form stable parallel straight lines within the time interval of 0 to 1 s. This observation suggests that the publishing frequency of these topics is highly stable, with negligible fluctuations in the time intervals between consecutive messages~\cite{zhu2024data}.

The numerical data further corroborate this finding. The four joint topics exhibit high message counts of 203, 203, 204, and 201, a mean average time interval of 5.00 ms, and a remarkably low standard deviation ranging from 0.01 to 0.16 ms. These statistics indicate that ROS is capable of precisely controlling the time intervals of message publishing when synchronizing these joint topics. Consequently, messages related to each joint can be reliably published at the anticipated frequency, thus ensuring high-quality synchronization, as detailed in Table~\ref{table:my_table}.

The timing diagrams of the color camera topics reveal relatively regular timestamp sequences and stable message publishing frequencies. According to the numerical table, the average time intervals of these topics range from 33.32 to 33.43 ms. Although the standard deviations, such as 1.88 ms, 1.90 ms, and 1.93 ms, are somewhat larger than those of the joint topics, they remain within an acceptable range for camera topics. This suggests that ROS can effectively synchronize the release of color camera images~\cite{hellmund2016robot}, allowing messages on these topics to be published at intervals close to the expected values.

For the depth-camera topics, the timing diagrams are also relatively stable, with average time intervals of 32.55 ms and 32.59 ms being relatively close. However, the standard deviations of 4.32 ms and 4.21 ms are somewhat larger than those of the color camera topics, although they still maintain a certain level of synchronization. In particular, the depth camera topic exhibits a relatively large average time interval of 34.93 ms and a standard deviation of 8.80 ms. This discrepancy may be attributed to interference factors encountered during the actual operation of the depth camera or its inherent acquisition and processing mechanisms, which could lead to slightly poorer synchronization. However, the overall synchronization quality of the majority of camera topics remains relatively good under ROS management, as illustrated in Fig.~\ref{fig:shijianchuo}.

\paragraph {\textbf{Stability of topic transmission frequency of ROS}}

\begin{figure*}  
  \centering          
  \includegraphics[width=6.5in]{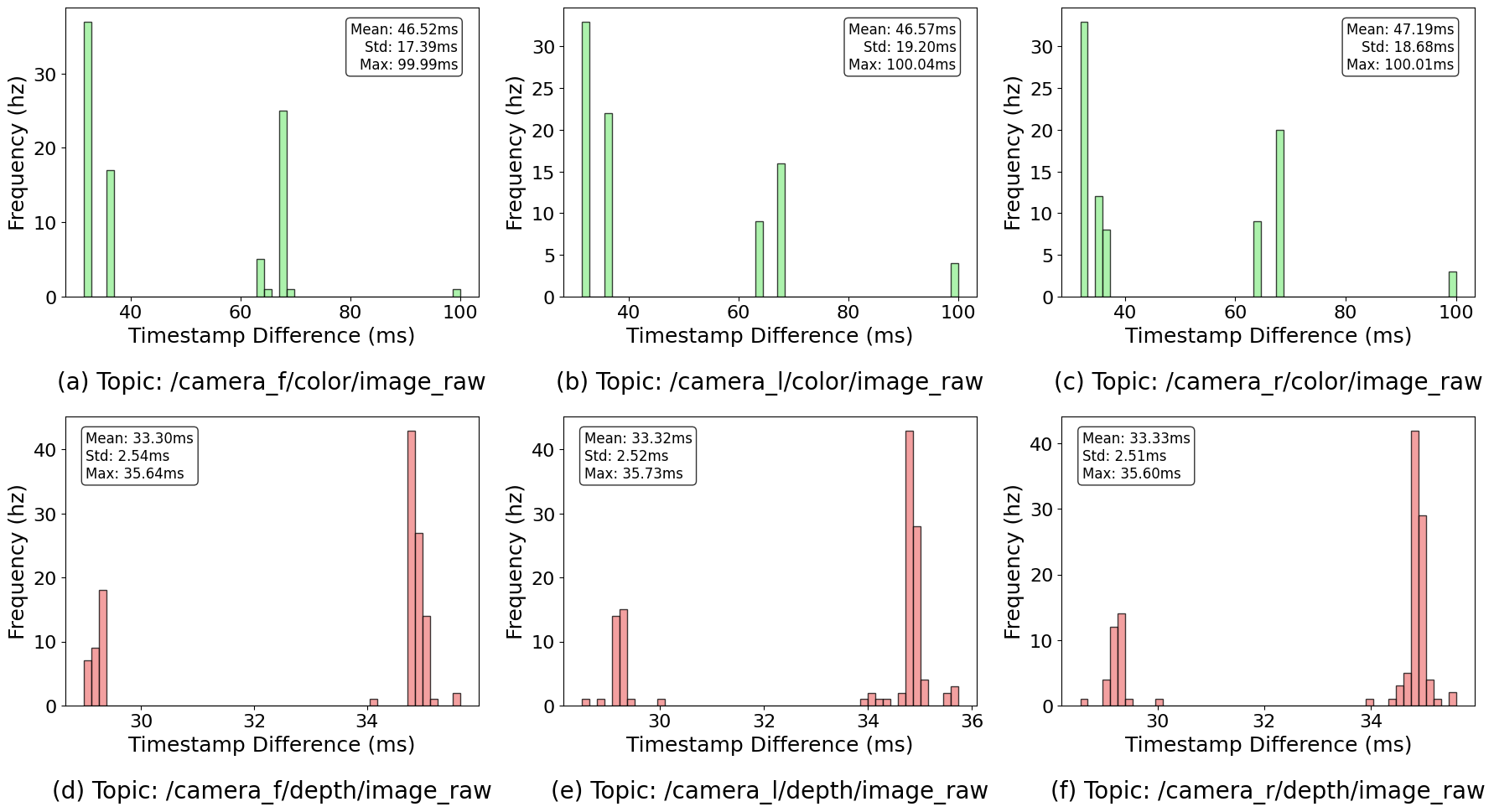}  
  \caption{Frequency of sending topics on the joint status of robotic cameras}  
  \label{fig:xiangji} 
\end{figure*}
\begin{figure*}  
  \centering          
  \includegraphics[width=5.5in]{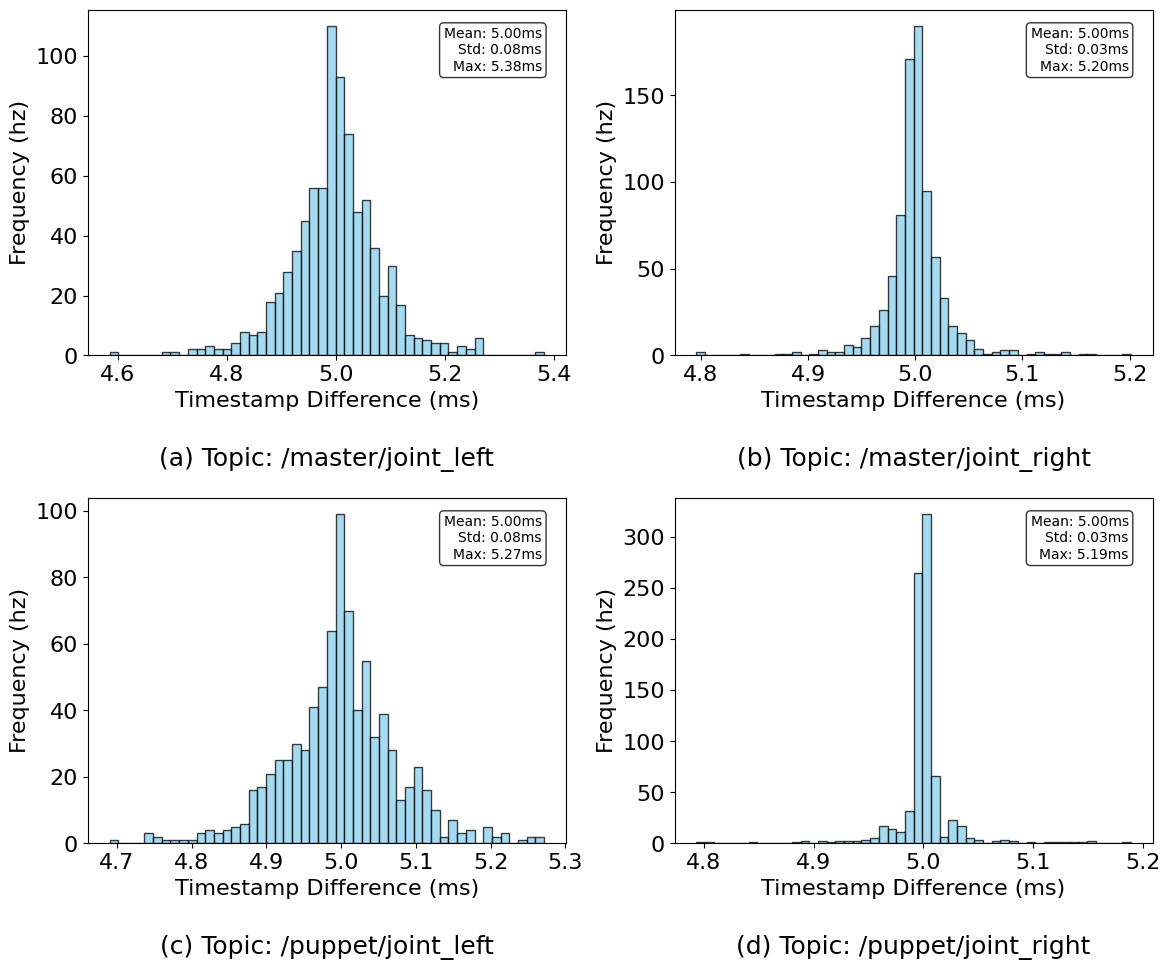}  
  \caption{Frequency of sending topics on the joint status of robotic arms}  
  \label{fig:guanjie} 
\end{figure*}

\begin{figure*}  
  \centering          
  \includegraphics[width=4.5in]{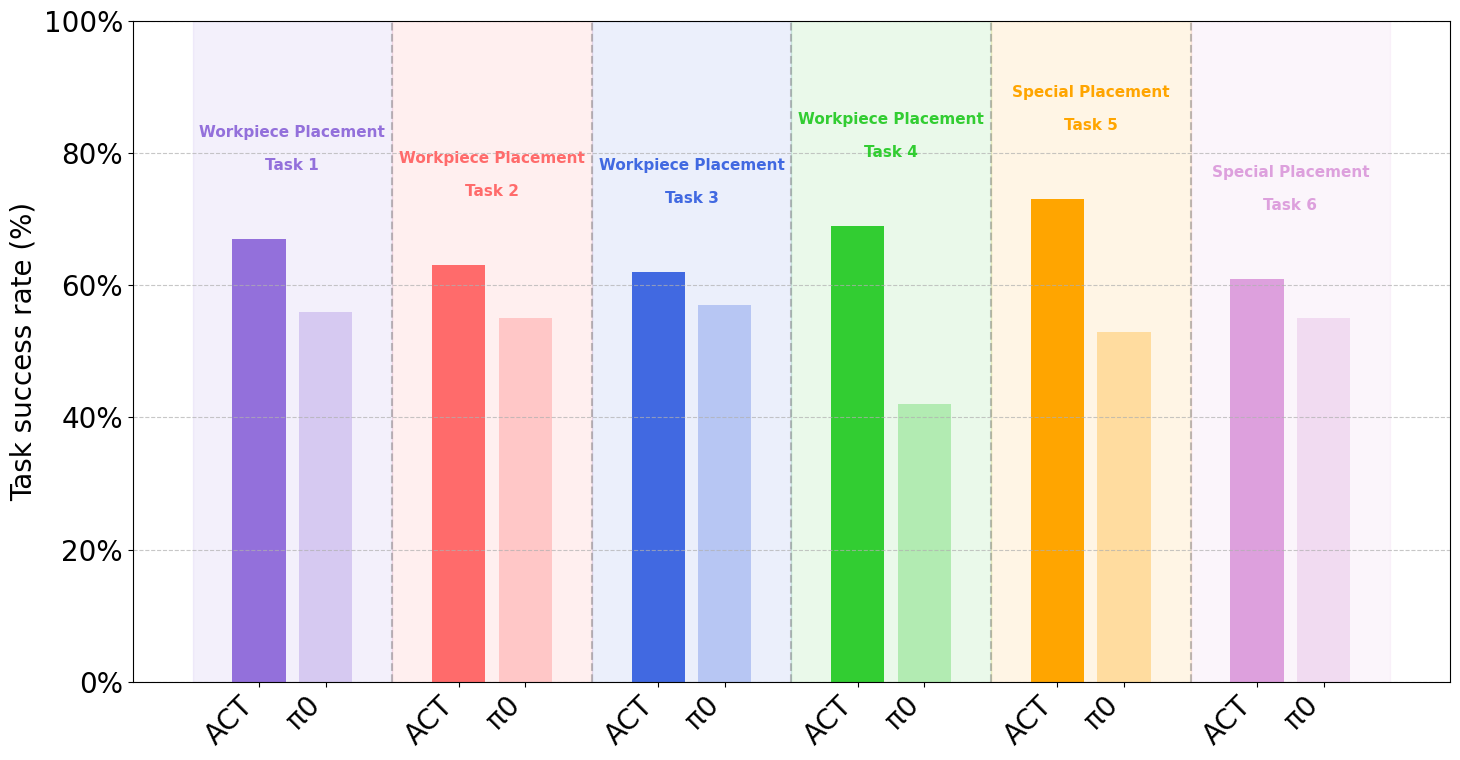}  
  \caption{Comparison of task success rates for different task types in ACT/$\pi$0 model}  
  \label{fig:模型对比} 
  
\end{figure*}
A detailed numerical analysis of ROS topics reveals distinct stability characteristics between different types of data streams. Specifically, JointState topics exhibit submillisecond precision with a mean interval of 5.000±0.010 ms, a remarkably low standard deviation of 0.007 ms, and a maximum latency of 5.099 ms (0.19\% deviation from the mean). These results meet the stringent industrial control requirements for cycle consistency.

In contrast, image topics display higher variability primarily due to constraints imposed by the large volume of data. The mean intervals for color and depth image topics are 46.523 ms and 46.573 ms, respectively, with standard deviations of 17.389 ms and 2.541 ms. The peak latencies for these topics are capped at 100.038 ms. Despite a coefficient of variation of 41\% for color images compared to 1.4\% for JointState data, temporal outliers are notably absent, since all maximum values are 100 ms. This confirms that the system maintains stability within the operational tolerances required for vision-based applications~\cite{breiling2017secure}. As illustrated in Figs.~\ref{fig:xiangji} and~\ref{fig:guanjie}.

\subsection{Model validation}

ACT in imitation learning is an end-to-end robot control algorithm that leverages the Transformer architecture~\cite{zhao2023learning}. This method segments a continuous action sequence into fixed-length blocks, such as 100 time steps at a frequency of 100 Hz. Using a multi-head attention mechanism, ACT integrates visual input (RGB images) with ontological perception (joint position and force feedback). It aligns the distribution of expert demonstration data using imitation learning objective functions such as cross-entropy loss and Kullback-Leibler divergence to generate coherent action sequences. The core innovations of ACT include reducing task complexity through action segmentation, introducing a time-integrated weighted average to mitigate the accumulation of composite errors, and incorporating comparative learning to enhance the robustness of state representation. These advances enable robots to efficiently imitate complex human operations such as opening cup covers and inserting batteries.

The $\pi$0 model within the vision-language-action (VLA) algorithm is an end-to-end robot control framework developed by the Physical Intelligence team~\cite{black2410pi0}. This model integrates vision, language, and action signals to perform common operational tasks. It features a three-stage architecture that combines a hybrid expert network with flow-matching technology. Using 10,000 hours of multimodal data pretraining, the model supports zero-shot semantic understanding and complex task decomposition. It is capable of generating high-precision action sequences in real time, such as cleaning new kitchens or completing industrial assembly tasks. Compared with traditional methods, the task success rate is significantly improved. Additionally, the model demonstrates adaptability to a variety of robotic arm hardware platforms. However, it relies on large-scale real-world data acquisition and requires fine-tuning and optimization for cross-platform migration. In general, the $\pi0$ model provides a closed-loop solution from perception to execution, advancing the capabilities of embodied intelligence.

The dataset facilitates collaborative optimization of models on different scales through a multitask hybrid arrangement comprising 4 basic tasks, 2 complex tasks, and a dynamic difficulty gradient design. For the small model ACT, the task success rate exceeds 60\%, particularly in dynamic disturbance scenarios, Tasks 5 and 6, highlighting the agility of small models. For the large model $\pi$0, the dataset effectively mitigates overfitting tendencies through high-precision annotations, such as spatial coordinates at the pixel level and long task chains, with the longest task containing 160K substeps, achieving a task success rate greater than 40\%. Experimental data show that models trained with this dataset exhibit high task success rates regardless of the size of the model. This underscores the universality of the data set across model scales and establishes it as a unified training benchmark for both lightweight and heavy-duty industrial robot algorithm deployments, as illustrated in Fig.~\ref{fig:模型对比}.

\section{Conclusion}

The Liaohe-CobotMagic-PnP dataset represents a crucial data infrastructure for industrial robot simulation, offering synchronous multimodal sensing, industrial-grade disturbance simulation, and coordinated dual-arm teleoperation. Built on industrial-grade multimodal interference modeling, the dataset demonstrates substantial applicability in dynamic manufacturing settings. Empirical evidence indicates that it can foster algorithm innovation for industrial robot systems by enhancing robustness and adaptability in real-world operational scenarios. This resource provides a valuable benchmark for advancing intelligent automation solutions within the Industry 4.0 context.

\bibliographystyle{IEEEtran}
\bibliography{IEEEabrv,dataset_publish}
\end{document}